%% file: seggan.tex
\documentclass[10pt,twocolumn,letterpaper]{article}

\usepackage{cvpr}
\usepackage{times}
\usepackage{epsfig}
\usepackage{graphicx}
\usepackage{amsmath}
\usepackage{amssymb}

\usepackage[pagebackref=true,breaklinks=true,letterpaper=true,colorlinks,bookmarks=false]{hyperref}

\usepackage{color}

\newcommand{\sys}{\mbox{SeGAN}}

\cvprfinalcopy 

\def\cvprPaperID{1561} 

\ifcvprfinal\fi
\begin{document}

\title{\sys : Segmenting and Generating the Invisible}

\author{Kiana Ehsani$^1$, Roozbeh Mottaghi$^2$, Ali Farhadi$^{1,2}$\\
$^1$ University of Washington, $^2$ Allen Institute for AI (AI2)
}

\maketitle
\thispagestyle{empty}

\begin{abstract}
Objects often occlude each other in scenes; Inferring their appearance beyond their visible parts plays an important role in scene understanding, depth estimation, object interaction and manipulation. In this paper, we study the challenging problem of completing the appearance of occluded objects. Doing so requires knowing which pixels to paint (segmenting the invisible parts of objects) and  what color to paint them (generating the invisible parts). Our proposed novel solution, \sys, jointly optimizes for both segmentation and generation of the invisible parts of objects. Our experimental results show that: (a) \sys \ can learn to generate the appearance of the occluded parts of objects; (b) \sys \ outperforms state-of-the-art segmentation baselines for the invisible parts of objects; (c) trained on synthetic photo realistic images, \sys \ can reliably segment natural images; (d) by reasoning about occluder-occludee relations, our method can infer depth layering. Code is available at \href{https://github.com/ehsanik/SeGAN}{https://github.com/ehsanik/SeGAN}.

\end{abstract}

\section{Introduction}

\input{intro2.tex}

\section{Related Works}

There is a large body of work on object detection \cite{RCNN, fastrcnn, he2014spatial, yolo, fasterrcnn, szegedy2014scalable}, semantic segmentation \cite{long2015fully, chen2016deeplab, chen2015attention, lin2015efficient, papandreou2015weakly, zheng2015conditional, dai2015boxsup, noh2015learning, liu2015semantic, krahenbuhl2015learning} and instance segmentation \cite{pinheiro2015learning, li2015iterative, dai2015instance, dai2016instance, pinheiro2016learning, zagoruyko2016multipath,ZhangCVPR16} using deep learning. These methods are designed for the visible regions of objects and they are not able to capture occlusions or provide a depth ordering for objects in an image. In contrast, our goal is to reconstruct occluded regions.

Occlusion reasoning has been studied in the literature extensively. \cite{winn06} propose a CRF for segmenting partially occluded objects. \cite{russell09} infer occlusion edges of polygons that represent objects. \cite{Gao2011ASO} make DPM more robust to occlusion by inferring whether a cell inside the object bounding box belongs to the object or not. \cite{guo2012beyond} use scene priors to infer the label for the occluded background regions. \cite{yang12} propose a layered object detection and segmentation method, where the goal is to infer depth ordering for the detected objects. \cite{hsiao12} propose an occlusion model for object instance detection based on 3D interaction of objects. \cite{girshick11,ghiasi14} propose methods for detection and pose estimation of occluded people. \cite{Pepikj_2013_CVPR} learn occluder-occludee patterns to improve object detectors. \cite{isola13} synthesize scenes by retrieving segments from training images, which requires reasoning about depth layers in the scene. \cite{tighe14} provide a semantic label for each pixel in an image along with the occlusion ordering for objects. \cite{chen15} use top-down information to tackle occlusions in multi-instance segmentation. We differ from all of these methods in that we complete the segmentation mask for the occluded objects and generate the appearance for the occluded regions of each object instance. Also, we show transfer from synthetic to natural images. 

The problem of bounding box completion has been tackled by \cite{amodalKarTCM15}, where the goal is to find the full extent of the object bounding box. Amodal segmentation methods have been proposed by \cite{zhu2015semantic, li2016amodal}, where they aim to provide a complete mask for occluded objects. The annotations that \cite{zhu2015semantic} provide is mainly based on the subjective judgment of the annotators (since the occluded parts of objects are not visible). In contrast, we modify our scenes by removing occluders and obtain an accurate groundtruth mask and texture for the occluded objects. The groundtruth annotation of \cite{li2016amodal} is obtained by pasting an object over an arbitrary image. Our argument is that occlusion relationships are not arbitrary and follow certain characteristics, and the way that we collect our occlusion data enables us to better model the occlusion relationships. Also, in contrast to these methods, we generate the appearance for the occluded regions.  

Conditional Generative Adversarial Networks (cGANs) \cite{mirza14} have been used for different applications such as prediction of future frames \cite{mathieu16}, style transfer \cite{li16}, colorizing and synthesizing images from edge maps \cite{isola16}, etc. Image inpainting using cGANs and DCGANs has been explored by \cite{pathak16} and \cite{yeh2016semantic}. In this paper, we combine cGANs with a convolutional network to segment and paint the occluded regions of objects simultaneously. Our problem is different from inpainting since our goal is to paint regions outside the input mask.

Recently, \cite{chen17} proposed a regression-based approach to synthesize images from a given semantic segmentation layout. Our method differ from \cite{chen17} since their goal is not to reconstruct occluded regions. Also, our method performs both segmentation and painting and it is category-agnostic.

\input{model3}

\section{Dataset}
\label{sec:dataset}
\begin{figure}[t]
\begin{center}
\includegraphics[width=20pc]{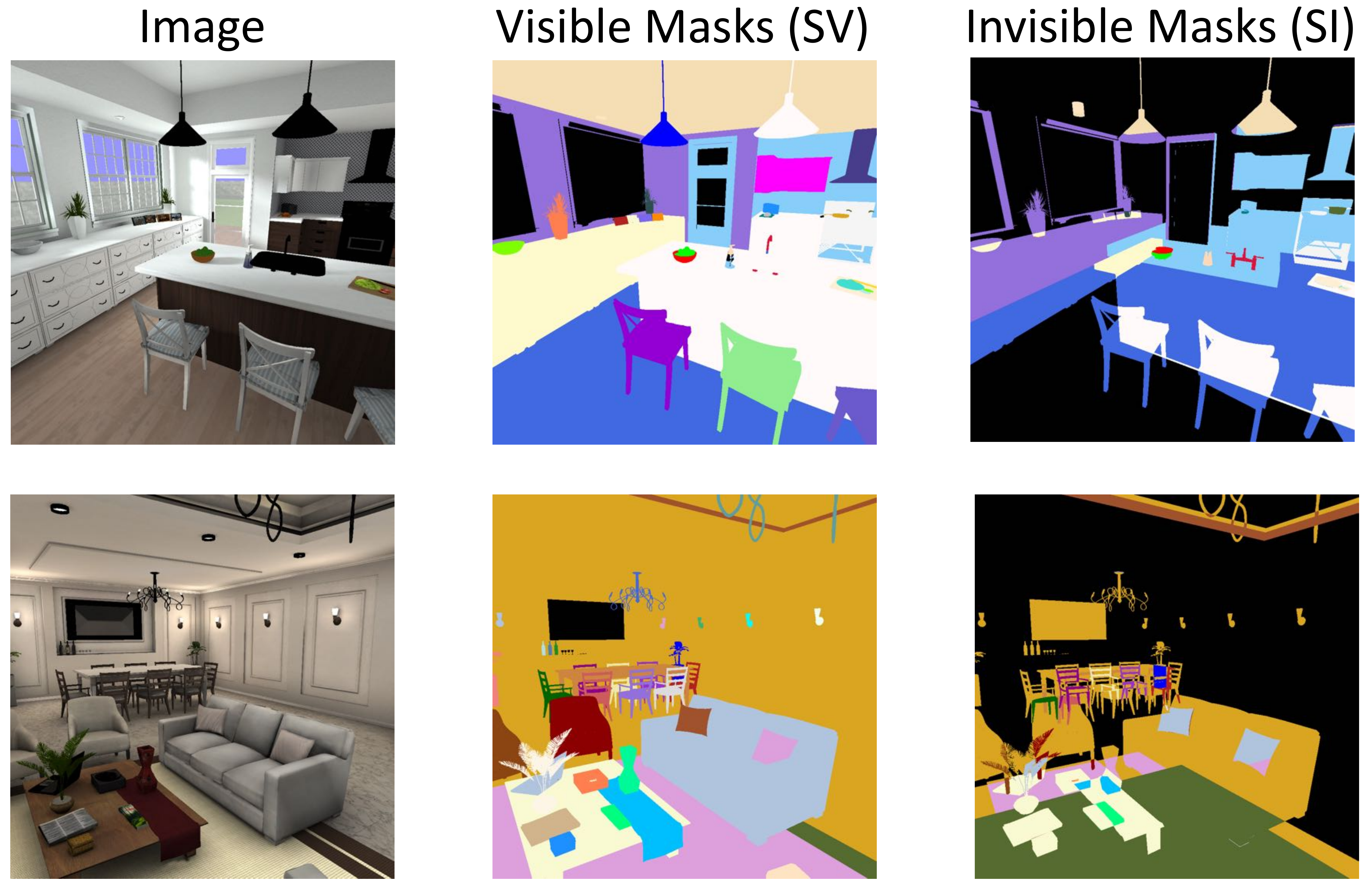}
\caption{\textbf{Example images of the dataset.} The first column shows the images captured from 3D synthetic scenes. The second column shows the segmentation mask for the visible regions. Each instance is encoded by a different color. The third column shows the invisible regions. For example, in the second row, the cushions occlude the sofa. Therefore, the regions behind the cushions have grey color in the third column, which means that those pixels belong to the grey sofa in the second column.}
\label{fig:dataset}
\end{center}
\end{figure}

In this paper, we introduce DYCE, a dataset of synthetic occluded objects. This is a synthetic dataset with photo-realistic images and natural configuration of objects in scenes. All of the images of this dataset are taken in indoor scenes. The annotations for each image contain the segmentation mask for the visible and invisible regions of objects. The images are obtained by taking snapshots from our 3D synthetic scenes. A few examples of images and their annotations are shown in Figure~\ref{fig:dataset}.

There are two advantages of a synthetic 3D dataset. First, we can obtain a 2D dataset of the desired size, and there is no restriction over the number of training samples we can generate. Second, we can move the camera to any location to capture interesting patterns of occlusion. We use the scenes of \cite{zhu17} to generate our dataset. 

\subsection{Generating 2D Images from 3D Scenes}
For generating the images, we change the location and the viewpoint of the camera in order to get a variety of images. For each scene, we generate 500 images from different viewpoints of the camera. We restrict the areas that the camera can be located. We move the camera in locations that the head of a person can be located in order to obtain common patterns of occlusion that people observe. We also restrict the orientation of the camera such that the camera points to objects in the scenes. Otherwise, the dataset will contain many images with no objects (for example, images depicting a portion of a wall). 

The procedure for generating the segmentation mask for the visible and invisible regions of objects is as follows. For each object, we generate an image with all other objects removed. Then, we compare this image with the original image, where no object is removed from the scene. The pixels that are the same in both images are the visible pixels of this particular object. To obtain the mask for the invisible region, we subtract the mask of the visible region from the mask of the full object.

\subsection{Statistics}
The number of the synthetic scenes that we use is 11, where we use 7 scenes for training and validation, and 4 scenes for testing.  Overall there are 5 living rooms and 6 kitchens, where 2 living rooms and 2 kitchen are used for testing. On average, each scene contains 60 objects and the number of visible objects per image is 17.5 (by visible we mean having at least 10 visible pixels). There is no common object instance in train and test scenes.

\input{experiment3}

\section{Conclusion}

In this paper, we address the problem of segmentation and appearance generation for the invisible regions of objects. We introduced \sys, which is a Generative Adversarial Network that generates the appearance and segmentation mask for invisible and visible regions of objects. Getting large-scale and accurate training data for this task is challenging. Our solution is to use photo-realistic synthetic data where we can obtain the exact boundaries of the invisible regions. Our experimental evaluations show that our model outperforms segmentation baselines, while it generates the appearance (as opposed to a binary mask). We also showed that our method outperforms GAN-based baselines for appearance generation and painting. We show generalization to natural images when the method is trained on synthetic scenes. Moreover, we evaluate our model for the task of depth layering and show improvements over a single image depth estimation baseline.

{\small
\bibliographystyle{ieee}
\bibliography{egbib}
}

\end{document}

%% file: intro2.tex
\begin{figure}
    \centering
    \includegraphics[scale=0.6]{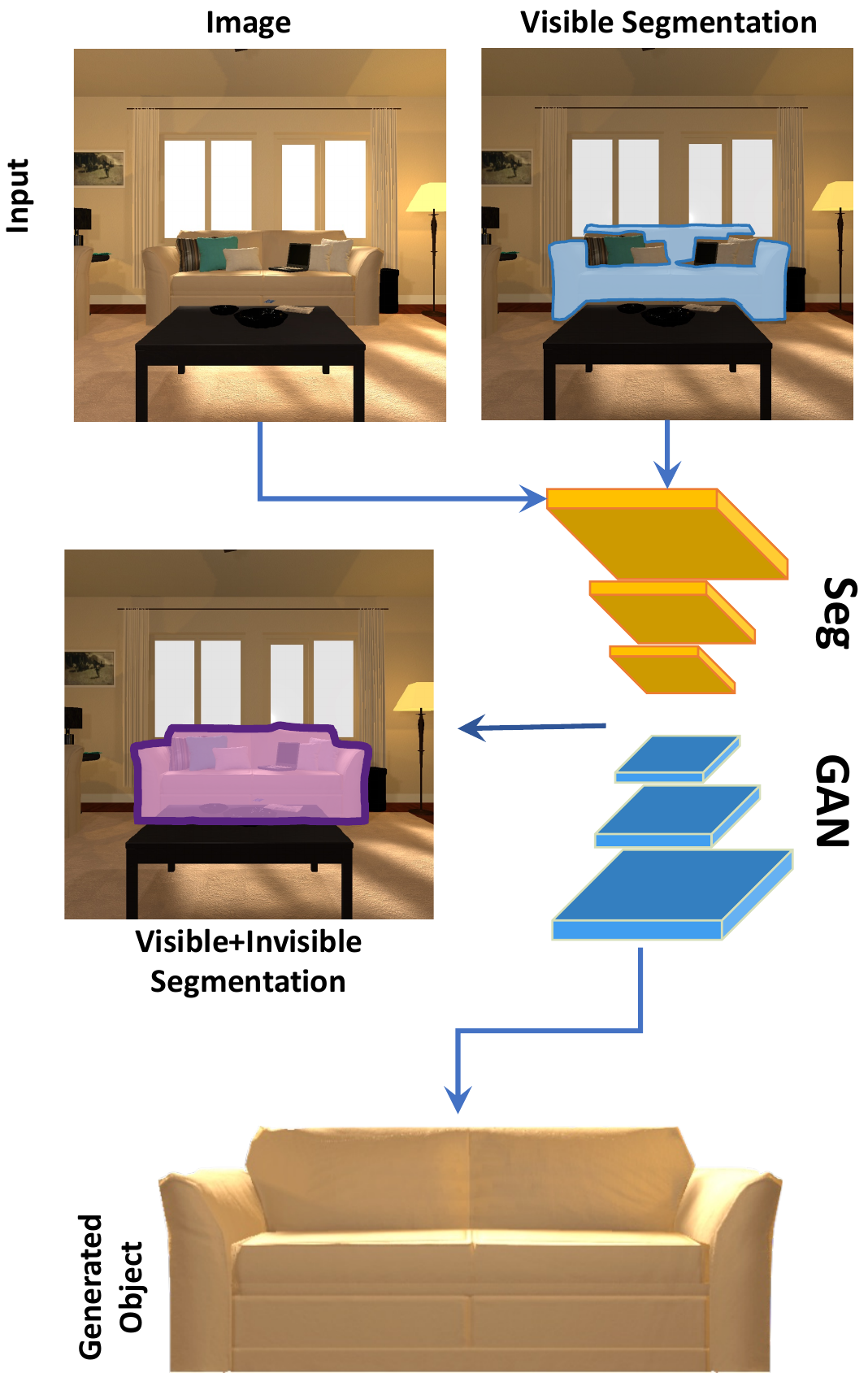}
    \caption{Our goal is to jointly segment and paint the invisible regions of objects. For instance, we predict how the sofa looks like when the occluders (cushions, laptop and coffee table) are removed. The input to our model is an image and a mask for the visible regions of objects (shown in blue).}
    \label{fig:teaser}
\end{figure}

Humans have strong ability to make inferences about the appearance of the invisible and occluded parts of scenes~\cite{aguiar02,newcombe99}. For example, when we look at the scene depicted in Figure~\ref{fig:teaser} we can make predictions about what is behind the coffee table, and can even complete the sofa based on the visible parts of the sofa, the coffee table, and what we know in general about  sofas and coffee tables and how they occlude each other. 
Devising algorithms to infer the appearance of what is behind an object exhibits several challenges. Predicting the appearance of the occluded regions of objects requires reasoning over multiple intertwined cues. Recognizing if an object is occluded or not is the first challenge to begin with. Second, knowing what pixels to color requires extending the boundaries of objects from their visible regions to invisible parts which requires some form of knowledge about the shapes of objects. The complex relations between the appearance of objects and the change in viewpoint and occlusion patterns form the third challenge.  Deformable objects can even make the problem ill-defined. Fourth, it is challenging to provide large-scale, accurate, and reliable training data to train models for occlusion reasoning. 

In this paper, we study the problem of generating the invisible parts of objects. Doing so requires segmenting the invisible part of the object and then generating the appearance of (painting) it. Inspired by the principles of learning-the-easy-thing-first, we propose \sys, a novel model that combines segmentation and generation and jointly optimizes for both of them. More specifically, given an input image and a segmentation for the visible regions of an object, our proposed GAN-based model learns to predict a segmentation for the occluded regions and generate the appearance by painting the invisible parts. Using segmentation masks of the invisible part as our intermediate step enables our network to learn about what pixels to paint before painting them. The generator network then paints the selected pixels. By jointly learning segmentation and generation networks \sys \  learns about the interdependencies between objects, their occlusion patterns, the shape and appearance of object segments. This allows us to address the first three challenges.

The key remaining challenge is training data; where can we find large-scale and accurate training data for what is behind the visible part of images? We argue that the proposed solution for Amodal segmentation in \cite{zhu2015semantic} is not suitable for our approach. Human judgements for predictions about the invisible parts of objects is subjective. Also, superimposing segments of images over other images~\cite{li2016amodal} would result in unnatural occlusion boundaries. In this paper, we propose to use photo-realistic synthetic data to learn how to extend segmentation masks from the visible parts of objects to the invisible regions and how to generate the appearance of the invisible part. Doing so allows us to obtain large-scale and accurate training data for the invisible regions of objects in images. 

Our experiments show that \sys \ can, in fact, segment and generate the invisible regions of objects.  Our results also show that our proposed segmentation network can learn to segment the occluded regions of objects and outperforms various state of the art segmentation baselines. We also show that our segmentation network can reliably segment the invisible parts of objects in natural images, when trained on our photo-realistic training set. By reasoning about occlusion patterns, our model can also make predictions about occluder-occludee relationships resulting in depth ordering inferences. Note that \sys \ is category-agnostic and does not require semantic category information. 

%% file: model3.tex
\section{Model}
\label{section:model}
\begin{figure*}[tp]
\centering
\includegraphics[width=40pc]{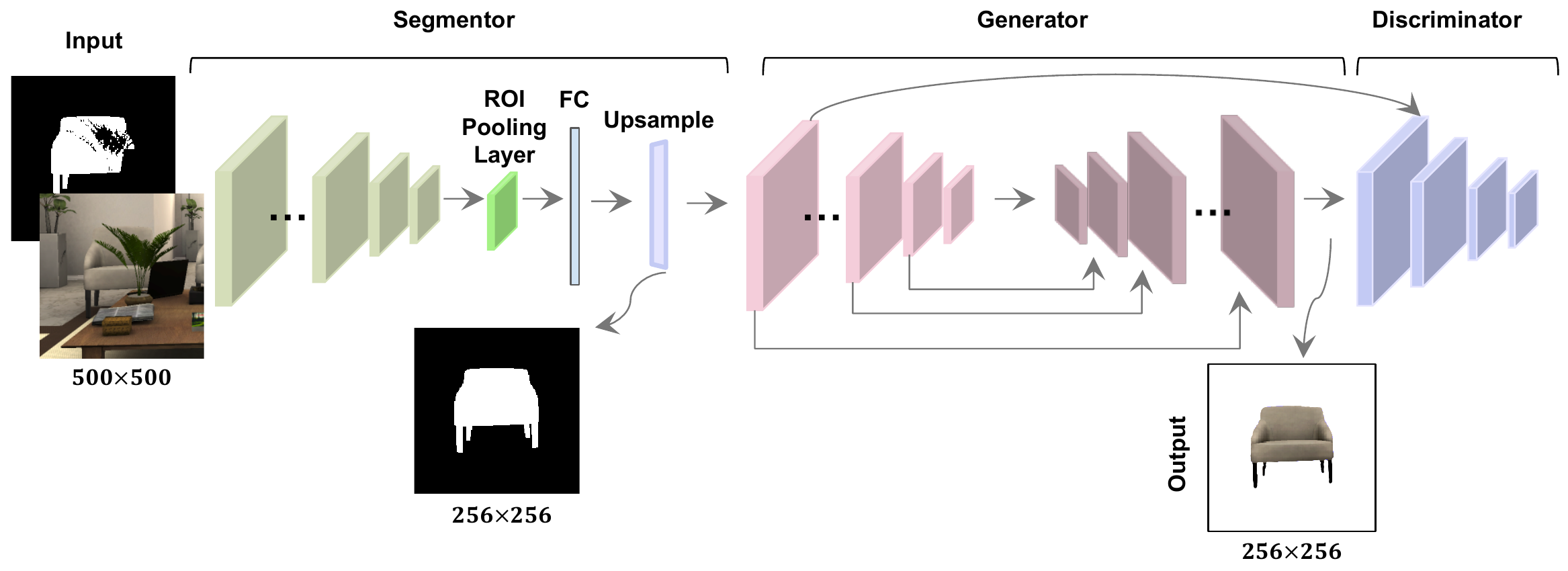}
\caption{\textbf{Model architecture.} Our network has three parts: \emph{segmentor}, \emph{generator}, and \emph{discriminator}. The input to our model is an RGB image and a mask for the visible region of an object which is obtained automatically by \cite{zagoruyko2016multipath}. The output is an RGB image that shows the appearance and segmentation for the full object (visible and reconstructed invisible regions). The segmentor part outputs an intermediate mask (the mask shown in the middle) that represents the full object, which is passed to the generator part of the network.}
\label{fig:model}
\end{figure*}
Our goal is to segment and paint the occluded regions of objects. The inputs to our model are a segmentation mask for the visible (non-occluded) regions of an object and an RGB image. The output is an RGB image where the occluded regions of that object have been reconstructed. The segmentation masks for visible regions can be obtained automatically from any instance segmentation method (e.g., \cite{zagoruyko2016multipath}).

We introduce \sys  \ that infers the mask for the occluded regions and paints those regions in a joint fashion. Our model has two main parts: (1) segmentation and (2) painting. The \emph{segmentation} part provides a mask for the occluded and non-occluded regions of objects, which is fed into the painting part of the model. The \emph{painting} part generates the appearance for the occluded region of the object. These two parts of the network are trained jointly. The architecture of the model is shown in Figure~\ref{fig:model}.

The segmentation part of the network is a CNN that takes a four-channel tensor as input, where three channels correspond to the RGB image, and there is a single channel for the segmentation mask of the visible region of an object. The mask for the visible region is obtained automatically (refer to Section~\ref{sec:exp} for details). The idea is to use the information from visible regions to segment and paint the invisible regions. We modify ResNet-18 \cite{he2015deep} to generate a mask image as output (the output of the last convolutional layer). Then, the mask output is fed into an ROI pooling layer. The ROI pooling layer is followed by a fully connected layer with the output size of 3364 ($58\times58$), and refer to its output by $o$. An upsampling layer converts $o$ to $256\times256$. We denote the output of the upsampling layer by $O$. Our final result is more accurate when we use upsampling.

The painting part of the network generates the invisible or occluded regions of the object. This part is a conditional generative adversarial network (cGAN) \cite{mirza14}, which consists of a generator and a discriminator.

The input to the generator, $M$, is computed as follows:

\begin{multline}
M(I, O, V) = \\ I\odot V + R\odot(O-V) + B\odot(J-O-V),
\end{multline}
where $\odot$ is element-wise multiplication, $I$ and $V$ are the input RGB image and input binary visible mask (\texttt{SV}), $J$ is an all-one matrix of size $256\times 256$, and $R$ and $B$ are $256\times 256$ images, where their first and third channels are 1s, respectively, and the rest of their channels are 0s. All of the binary masks in the above equation are repeated three times to form a 3 channel image. Basically, in the generator's input, the mask for the invisible region (which is provided by the segmentation part of the network) is red, and the region outside the mask is blue. 

We adopt Unet~\cite{ronneberger2015u} for the generator network, which is an encoder-decoder with skip connections from encoders to the corresponding layers in the decoder. The discriminator network includes four convolutional layers, followed by one sigmoid layer. The architecture for this part is similar to that of the Pix2Pix network~\cite{isola16}.

The loss function for our model is a combination of the losses for segmentation and painting.
For segmentation, we define a customized loss function using binary cross entropy loss that is computed on the prediction of the network and the groundtruth for the full object binary mask (referred to as \texttt{SF}). In Section~\ref{sec:dataset}, we explain how we obtain accurate groundtruth for the occluded regions. Ideally, the segmentation part should learn 1) not to change the mask for the pixels in \texttt{SV} (mask for visible regions) and 2) to predict the mask for the pixels in \texttt{SI} (mask for invisible regions) correctly. The binary cross entropy loss is defined as: 
\begin{equation}
L_{ent}^S(g,o) = - \frac{1}{n} \sum_{ij\in S} (g_{ij} log(o_{ij}) + (1 - g_{ij}) log(1 - o_{ij})),
\end{equation}
where $S$ is a subset of pixels (e.g., pixels of the visible region), $g_{ij}$ and $o_{ij}$ are pixels at location $(i,j)$ of the groundtruth \texttt{SF} and predicted mask, respectively, and $n=|S|$.

Our loss function for segmentation is defined as:
\begin{equation}
L_{segm}(g,o) = \lambda_{bg} L_{ent}^{\overline{SF}}(g,o) + \lambda_{SV} L_{ent}^{SV}(g,o) +  \lambda_{SI} L_{ent}^{SI}(g,o),
\end{equation}
where $\overline{SF}$ is the set of pixels in the image patch not in $SF$, or in other words the pixels that do not belong to either visible or invisible parts of the object. A sigmoid function is applied to the predicted output so we obtain a real number between 0 and 1. The intuition for defining this objective is to differentiate among making mistakes in segmenting the visible region, invisible region and the background.

The loss function for painting is defined as follows:
\begin{multline}
L_{cGAN} (G, D) = E_{x\sim p_{data}(x),z\sim p_z(z)} [\log D(G(x, z))]\\+ E_{x\sim p_{data}(x),z\sim p_z(z)}
[\log(1 - D(x, G(x, z))],
\end{multline}
where $G$ and $D$ are the generator and the discriminator networks, respectively, $x$ is the input
and $z$ is a random Gaussian noise vector, which is mainly used for regularizing the generator network. 

Previous approaches found L1 and L2 distance losses to be helpful for GANs~\cite{pathak16, isola16}, thus the final loss function for the adversarial part is defined as:

\begin{equation}
L^*(G) =  L_{cGAN} (G, D) + \lambda L_{L1}(G)
\end{equation}

The loss function for our \sys \ end-to-end model, $L_{full}$, is defined as: 

\begin{equation}
L_{full} = \lambda_{L^*} L^*(G) + L_{segm}(g,o)  
\end{equation}

%% file: experiment3.tex
\section{Experiments}
\label{sec:exp}
Our model performs segmentation and painting jointly. Hence, in this section, we evaluate our model from these two perspectives. In addition, we show results of generalization to natural images. Finally, we present our evaluation for the depth layering task. Our training and test sets include 41924 and 27617 objects depicted in 3500 and 2000 images, respectively.

\subsection{Implementation details}

The segmentation part is initialized by the weights of ResNet-18 \cite{he2015deep} that are pre-trained on ImageNet \cite{imagenet_cvpr09}. We use random initialization for the painting part of the network.

All input images and their masks are resized to $500\times 500$. We used bilinear interpolation for resizing. Thus, the segmentation mask might contain values in the interval $(0,1)$. To obtain bounding boxes for the ROI pooling layer we expand the box around the input \texttt{SV} masks by a random ratio between 10-30\% from each side. Note that we ignore the portions that lie outside the image. We compute the segmentation loss on groundtruth segmentation masks of size $58\times58$ (for each object, we crop the image using the expanded bounding box and scale the cropped image to $58\times 58$). Then, we upsample the predicted mask to $256\times 256$ using a bilinear upsampling layer and use the $256\times256$ mask as the input to the painter network. We do not train the upsampling layer. The generator outputs a three channel $256\times 256$ image, which includes the RGB values for the full object (invisible and visible regions).

We use the following coefficients in the loss function: $\lambda_{bg}=1, \lambda_{SV}=5, \lambda_{SI}=3$, $\lambda_{L1}=100$, and $\lambda_{L^*}=0.1$. These values are obtained using a validation set. Also, to help the network to converge, we first train the segmentation network and the generator network jointly and then train the whole network end to end.

\begin{table*}[t]
\setlength{\tabcolsep}{3pt} 
\centering
\begin{tabular}{|c|c|c|c|c|c|c|}
\hline
Model  & Training Data & Input Mask & Loss Mask & Visible $\cup$ Invisible & Visible & Invisible \\ \hline
IBBE \cite{li2016amodal} & natural and synthetic & \texttt{SV} & \texttt{SF} & 31.0 & 25.2 & 5.1 \\ \hline
Multipath \cite{zagoruyko2016multipath} & natural and synthetic & \texttt{SV} & \texttt{SF} & 36.0 & 34.8 & 12.3      \\ 
\hline
Multipath \cite{zagoruyko2016multipath}  & natural &\texttt{SV} & \texttt{SV} & 47.51 & 48.58 & 6.01 \\
\hline
Pix2Pix \cite{isola16}  & natural and synthetic &\texttt{SV} & \texttt{SF} & 52.3 & 49.6 & 11.9 \\
\hline
\sys \ (ours) w/ predicted \texttt{SV}  & natural and synthetic & \texttt{SV} & \texttt{SF} & \textbf{68.78}   & \textbf{64.76}    & \textbf{ 15.59 }      \\ 
\hline
\hline
\sys \  (ours) w/ GT \texttt{SV} & synthetic & \texttt{SV} & \texttt{SF} & 75.71      &  68.05   &   23.26    \\ \hline
\end{tabular}
\caption{\textbf{Segmentation evaluation.} We compare our method with \cite{li2016amodal}, \cite{zagoruyko2016multipath}, and \cite{isola16} on the synthetic test data. \texttt{SV} and \texttt{SF} refer to the mask for visible regions of objects and the full object, respectively. We evaluate how well we predict `Visible' regions, `Invisible' regions and their combination. The bottom row is not comparable with other rows since it uses groundtruth information.}
\label{tab:synres}
\end{table*}

\subsection{Evaluation}

\noindent \textbf{Segmetation \& Painting.} We evaluate our model, \sys  , in two settings. First, we use the output of the Multipath network \cite{zagoruyko2016multipath}, which is a state-of-the-art model for generating the segmentation mask for objects as our input mask for the visible regions (\texttt{SV} masks). Secondly, to factor out the effects of \texttt{SV} segmentation approach from our results, we also show the results using the groundtruth mask as the input for the visible region of the object. 

 After obtaining segmentation masks from Multipath, we find the segmentation mask that corresponds to the visible region of the groundtruth training object. The segmentation mask that has the largest intersection over union with the visible region of the groundtruth mask is selected as the input mask during training. For evaluation, we consider all masks generated by Multipath.

\begin{figure*}
    \centering
    \includegraphics[width=35pc]{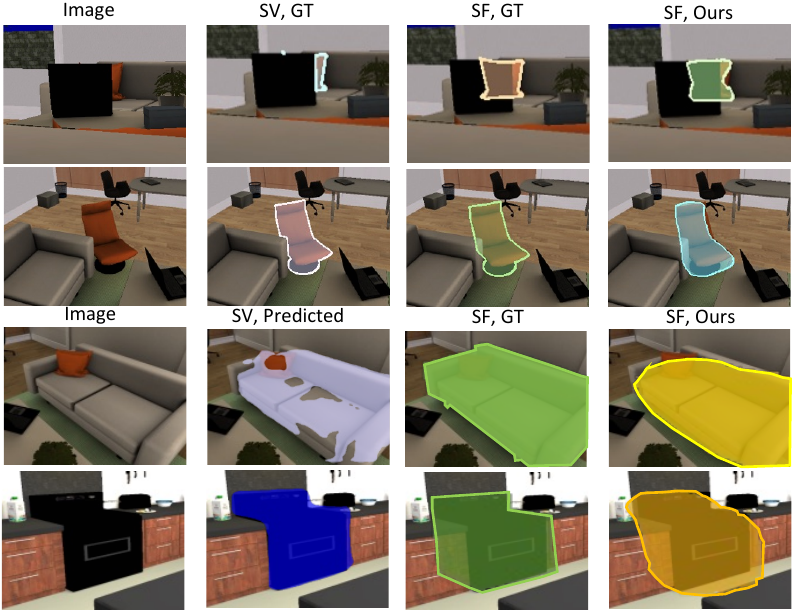}
    \caption{\textbf{Qualitative results of segmentation.} We show the results using groundtruth for the visible region (\texttt{SV}, GT) in the first two rows, and using the predicted mask in the last two rows. The groundtrouth for the full object (\texttt{SF}, GT), and our predicted mask for the full object (\texttt{SF}, Ours) are also shown.\vspace{-0.3cm}}
    \label{fig:synres}
\end{figure*}

We evaluate our model using three metrics for segmentation and two for painting. For segmentation, we evaluate how well we predict (1) the mask for the occluded regions (\texttt{SI}), (2) the mask for non-occluded regions (\texttt{SV}), and (3) the mask for the full object (\texttt{SF}=\texttt{SV} $\cup$ \texttt{SI}). The intuition for evaluating the mask for visible regions is to check whether our approach distorts the input mask when the object is not occluded. For all of these settings, we compute intersection over union between the predicted mask of the model and the groundtruth mask. For evaluating painting, we use L1 and L2 distance of the predicted output and the ground truth image. 

\begin{figure*}[t]
    \centering
    \includegraphics[width=40pc]{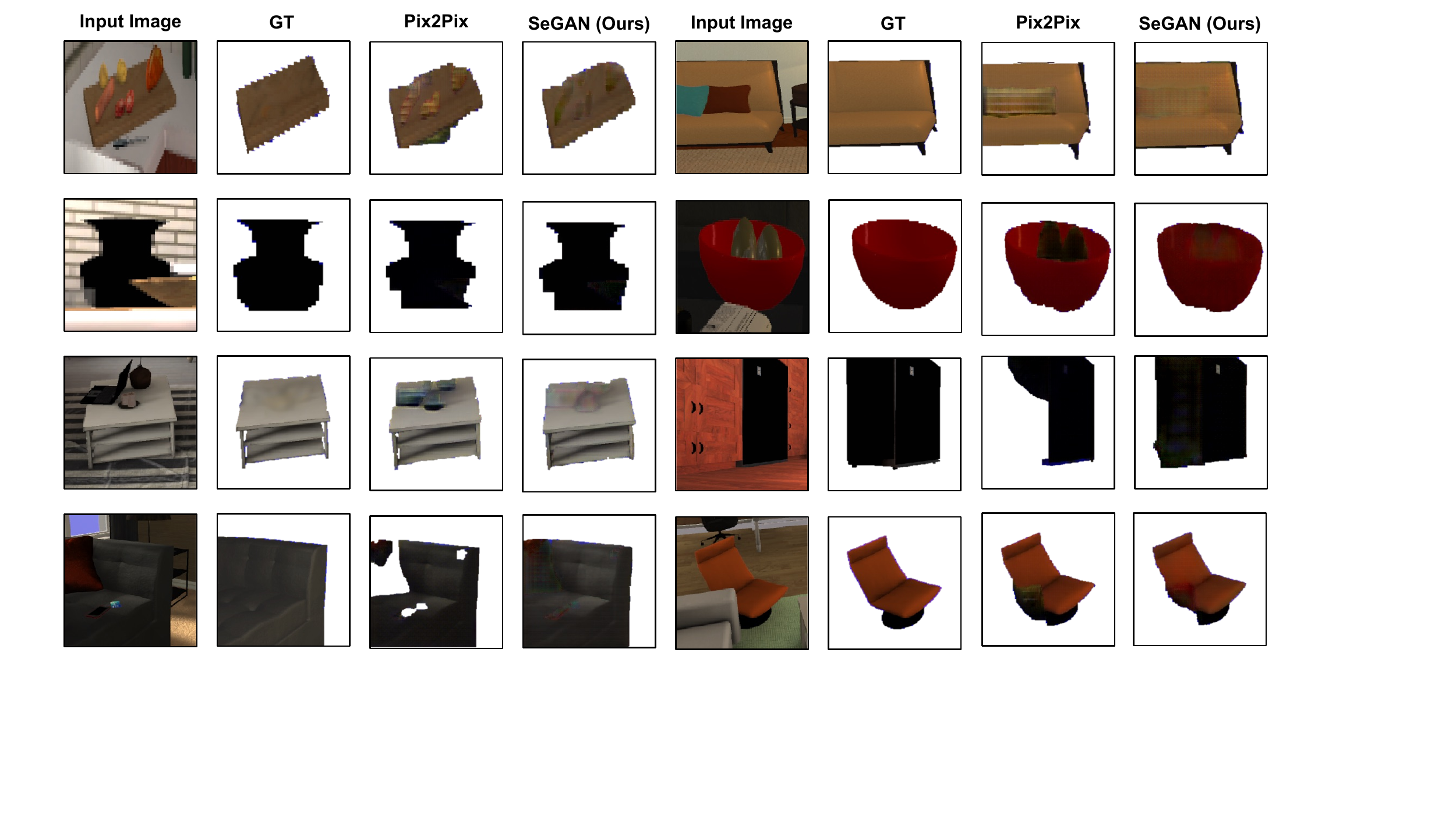}
    \caption{\textbf{Qualitative results of painting.} We show the input image, the groundtruth object (without any occlusion) and the result of Pix2Pix \cite{isola16} network.}
    \label{fig:texture}
\end{figure*}

Table~\ref{tab:synres} summarizes our results for segmentation. First, our method (referred to as `\sys \  w/ predicted SV') significantly outperforms Multipath for the task of predicting masks for the full object (\texttt{SF}) and invisible regions (\texttt{SI}). 
It is interesting to see that our method improves the segmentation of the visible regions (\texttt{SV}) as well. We have two variations of the Multipath network as our baselines: one trained only on natural images data (trained on MS COCO dataset \cite{lin14}) and one trained on the combination of natural images data and our synthetic images (trained to predict occluded and visible regions). Another baseline is IBBE \cite{li2016amodal} (an amodal segmentation method), which is fine-tuned on our synthetic data. We also compare with Pix2Pix \cite{isola16}, where it receives the same inputs as our model and generates the appearance for the full object. Using groundtruth masks (\sys \ w/ GT \texttt{SV}) shows that our method will perform even better if it receives more accurate masks for the visible regions as the input. Figure~\ref{fig:synres} shows qualitative segmentation results of our method.

We now evaluate our network on appearance generation (painting). Our first baseline is a nearest neighbor method. We feed the image into ResNet-18 pretrained on ImageNet and obtain features from the layer before the classification layer. Similarly, we feed the mask image for the visible region into ResNet and obtain features for the mask as well. We concatenate these features. For each test example, we find the training image and mask that has the smallest distance (L2 distance on the concatenated features). 

As another baseline, we use the Context-Encoder network~\cite{pathak16}. The main task of their network is to complete a cropped patch of an image using the context around it, so we crop a box around the object but leave the visible pixels unchanged. In other words, we remove all the pixels on the image patch that can potentially belong to the invisible regions and calculate the loss on the full object.

As our last baseline, we train the Pix2Pix network \cite{isola16} on our dataset by feeding just the pixels for the visible part of the image as an input and calculating the loss on the full object (visible and invisible regions). The network is supposed to learn to generate the appearance of the full object. 

Table~\ref{tab:texture} shows the results for painting. Our model outperforms all of the baselines methods. Again, for factoring out the performance of the \texttt{SV} prediction methods we repeat all of the experiments with groundtruth \texttt{SV} masks. The qualitative results for this experiment can be seen in Figure~\ref{fig:texture}.

\begin{table}[t]
\setlength{\tabcolsep}{3pt} 
\centering
\begin{tabular}{|c|c|c|c|c|c|c|}
\hline
Model  & Input & L1 & L2 \\ \hline
Nearest Neighbor (NN)  & Image + $SV_p$  & 0.20 & 0.12 \\ \hline
Context-Encoder \cite{pathak16}  & Image & 0.18 & 0.09 \\ \hline
Pix2Pix \cite{isola16} & Image + $SV_p$ & 0.15 & 0.09 \\ \hline
\sys \ (ours)  & Image + $SV_p$ & \textbf{0.07} & \textbf{0.03} \\ \hline \hline
\sys \ (ours) & Image + $SV_{gt}$ & 0.03 & 0.01 \\ \hline
\end{tabular}
\caption{\textbf{Painting evaluation.} We use L1 and L2 distances as the evaluation metric so lower numbers are better. $p$ and $gt$ subscripts refer to predicted masks (by \cite{zagoruyko2016multipath}) and groundtruth, respectively.\vspace{-0.3cm}}
\label{tab:texture}
\end{table}

We also performed a human study using Mechanical Turk. Table~\ref{tab:human} shows how often the result of a particular method is chosen as the best generated image.
\begin{table}[h]
\setlength{\tabcolsep}{3pt} 
\centering
\begin{tabular}{|c|c|c|c|}
\hline
Model  & NN & Pix2Pix \cite{isola16} & \sys \ (ours) \\ \hline
 Chosen as best &  0.46\% & 29.74\% & \textbf{69.78\%} \\ \hline 
\end{tabular}
\caption{\textbf{Human study results.} We show how often the subjects select the generated images of our method versus two baselines: nearest neighbor (NN) and Pix2Pix \cite{isola16}.}
\label{tab:human}
\end{table}

\noindent \textbf{Generalization to natural images.} We now evaluate whether our model generalizes to natural images when it is trained on our synthetic data. A major issue is that there is no dataset of natural images that provides large-scale and accurate mask annotations and texturing for both occluded and non-occluded regions of objects. \cite{zhu2015semantic,li2016amodal} construct datasets for object occlusion, but occlusion patterns of \cite{li2016amodal} are unnatural, and the dataset of \cite{zhu2015semantic} does not include the appearance for the occluded regions.  Therefore, we report the results only for the segmentation task.

\begin{figure}
    \centering
    \includegraphics[width=18pc]{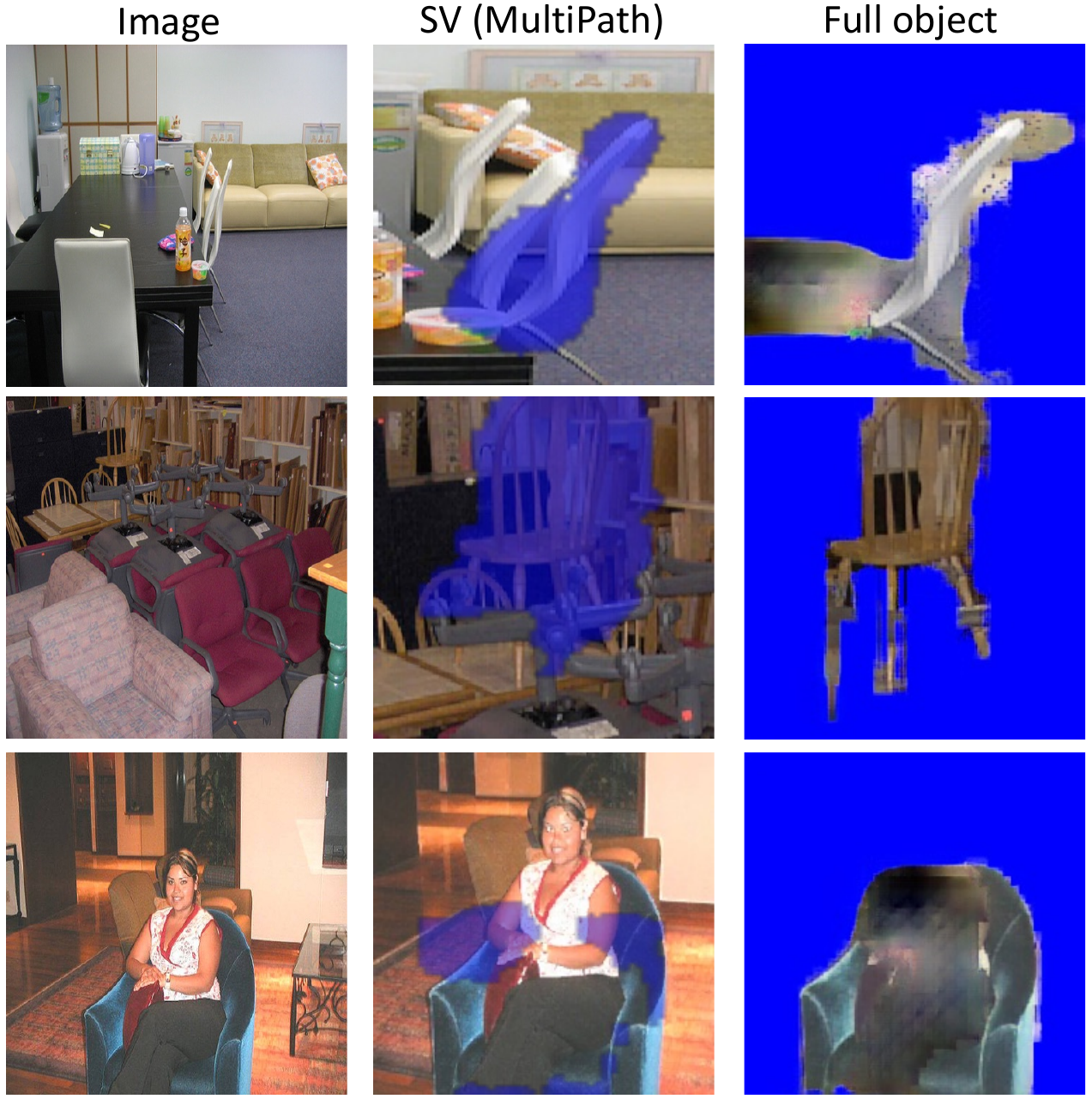}
    \caption{\textbf{Qualitative results of generalization to natural images.} We show the prediction for the visible region using Multipath (\texttt{SV}) and segmentation and painting results of our method.\vspace{-0.5cm}}
    \label{fig:natres}
\end{figure}

To evaluate the performance of our model on natural images, we construct a dataset using PASCAL 3D dataset \cite{xiang14}. PASCAL 3D associates a 3D CAD model to each object instance in the dataset and it also provides annotations in terms of azimuth, elevation and distance of the camera with respect to the objects. Therefore, we can project the 3D CAD model onto the image and obtain the segmentation mask for full objects. Note that the projection does not have any occlusion information so it generates the mask for the full object (\texttt{SF}) as if it is not occluded by any object. On the other hand, datasets such as \cite{Hariharan2011SemanticCF} provide the segmentation mask for the visible region of the objects. Hence, we can obtain the mask for the occluded regions by subtracting these masks from the mask we obtain by projecting the CAD models. We use five indoor categories of PASCAL 3D dataset (i.e. bottle, chair, diningtable, sofa, and tvmonitor) for our experiments. As before, we run Multipath on these natural images to obtain the segmentation masks for the visible regions.  

\begin{table}[t]
\setlength{\tabcolsep}{3pt} 
\centering
\begin{tabular}{|c|c|c|c|}
\hline
Model              & Vis. $\cup$ Invis. & Visible & Invisible \\ \hline
Multipath \cite{zagoruyko2016multipath}  (natural) & 46.8 & \textbf{58.9}    & 13.9      \\ 
\hline
Multipath \cite{zagoruyko2016multipath}  (natural+syn.) & 46.3 & 54.0  & 8.4  \\ 
\hline
Ours w/ predicted \texttt{SV} & \textbf{47.3}   & 58.1    & \textbf{18.7}      \\ 
\hline
\hline
Ours w/ GT \texttt{SV}       &  50.7         & 58.9 & 23.1         \\ \hline
\end{tabular}
\caption{\textbf{Evaluation of generalization to natural images.} The input masks of our network are \texttt{SV} masks, and the groundtruth mask for the loss function is \texttt{SF}.\vspace{-0.3cm}}
\label{tab:realres}
\end{table}

Table~\ref{tab:realres} shows the results of generalization to natural images. Although, our method only uses synthetic occlusion information for training, it is still able to provide accurate results on natural images. More importantly, our method outperforms Multipath on segmentation of the invisible regions (\texttt{SI}), while there is only a slight degradation in the performance for the visible regions. Combining natural and synthetic data makes the performance worse. It probably makes the training more difficult. Example predictions of our method on natural images are shown in Figure~\ref{fig:natres}.

\noindent \textbf{Depth layering.} The final experiment for evaluating our network is depth layering i.e. it infers the depth layers for a pair of objects with respect to the camera. Depth layering can be inferred from occlusion relationships since typically the occluder is in front of the occludee and is closer to the camera. The groundtruth data for this problem can be easily obtained from our dataset since we have access to the occlusion relationships in the data. 

For this task, we first predict the segmentation mask for the full object (\texttt{SF}) using our network. Then amongst the rest of the objects in this scene we find the ones, whose segmentation mask for visible region (\texttt{SV}) intersects with the predicted mask for the invisible region (\texttt{SI}). An object $q$ occludes object $p$ if intersection over union of the segmentation mask for the visible region (\texttt{SV}) of object $q$ with the segmentation mask of the invisible region (\texttt{SI}) of object $p$ is above a threshold. The threshold that we use is $5\%$. 

As the baseline for this task, we use the method of \cite{eigen2015predicting}, which infers depth map from a single image. To obtain the depth layering result for this method, for each image, we project the groundtruth segmentation mask for the visible region (\texttt{SV}) onto the output of the depth estimation and compute the average depth value for all the pixels that are inside the mask. This will be the estimated depth for this object. Then, for each pair of occluding objects we compare their depth, and the one that has a lower depth will be closer to the camera and, therefore, we consider it as the occluder. 

The evaluation metric is defined as the average ratio of number of correct predictions over the number of occlusion pairs for each image.

\begin{figure}
    \centering
    \includegraphics[width=20pc]{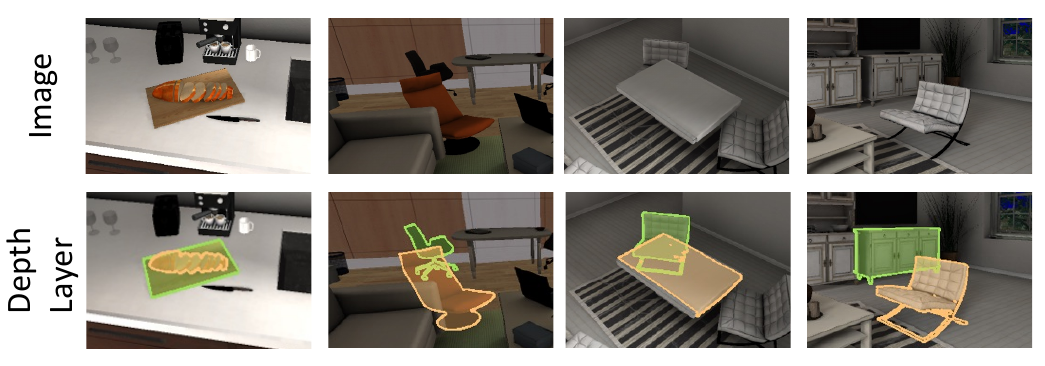}
    \caption{\textbf{Qualitative results of depth layering.} We predict that the object shown with the orange mask is closer to the camera than the object shown with the green mask.\vspace{-0.3cm}}
    \label{fig:depthres}
\end{figure}

We report the average over all images in the test dataset. The number of occlusion pairs in our test set is 4M pairs. The result for this task is shown in Table~\ref{tab:depthres}. This task has been evaluated on the synthetic data since we have accurate depth and occlusion information. Figure~\ref{fig:depthres} shows the qualitative results of depth layering. 

\begin{table}[t]
\centering
\begin{tabular}{|c|c|}
\hline
Single Image Depth \cite{eigen2015predicting} & Ours\\
\hline
 60.18 & \textbf{84.04}\\
\hline
\end{tabular}
\caption{\textbf{Depth layering results.} We compare our depth layering results with the results of \cite{eigen2015predicting}, which estimates depth from single RGB images. We modify \cite{eigen2015predicting} to use masks.\vspace{-0.3cm}}
\label{tab:depthres}
\end{table}

There are several reasons why our model is more accurate. First, in many cases the occluder and occludee do not differ enormously in depth, which makes it difficult for the baseline to infer the occlusion relationship. Second, the predicted depth map is computed over the entire image. Hence, many low-level details have been removed while our predicted mask is for a specific object and does not lose much information around the occlusion boundaries. Third, our network predicts the invisible regions of the object, but the depth estimator does not have this capability. 

